\documentclass[10pt,twocolumn,letterpaper]{article}

\usepackage{cvpr}              

\usepackage{booktabs, colortbl}
\definecolor{cvprbest}{RGB}{199,233,192}   
\definecolor{cvprsecond}{RGB}{228,238,188} 
\definecolor{cvprthird}{RGB}{255,247,174} 
\newcommand{\best}[1]{\cellcolor{cvprbest}\textbf{#1}}
\newcommand{\second}[1]{\cellcolor{cvprsecond}#1}
\newcommand{\third}[1]{\cellcolor{cvprthird}#1}
\usepackage{graphicx}
\usepackage{animate} 
\usepackage{caption}    
\usepackage{subcaption}  
\usepackage{float}

\usepackage[normalem]{ulem}

\definecolor{cvprblue}{rgb}{0.21,0.49,0.74}
\usepackage[pagebackref,breaklinks,colorlinks,allcolors=cvprblue]{hyperref}

\def\paperID{} 
\def\confName{CVPR}
\def\confYear{2026}

\title{AeroDGS: Physically Consistent Dynamic Gaussian Splatting for Single-Sequence Aerial 4D Reconstruction}

\author{
Hanyang Liu\\
The Ohio State University\\
{\tt\small liu.12021@buckeyemail.osu.edu}
\and
Rongjun Qin\\
The Ohio State University\\
{\tt\small qin.324@osu.edu}
}

\begin{document}

\twocolumn[{
\maketitle
\vspace{-0.6em}

\begingroup
\newlength{\teaserH}\setlength{\teaserH}{0.2\textwidth}
\newlength{\gap}\setlength{\gap}{0.2em} 

\begin{center}
  \begin{minipage}[t]{0.33\linewidth}\centering
    \vspace{4pt}
    \begin{minipage}[c][\teaserH][c]{\linewidth}\centering
      \animategraphics[autoplay,loop,poster=first,
        width=\linewidth,keepaspectratio]{15}{fig/original_frames/vidA_}{0001}{0016}
    \end{minipage}
    \par\vspace{2pt}\parbox[t]{\linewidth}{\centering \textbf{(a)}~Input aerial video captured from a moving flight over urban scenes}
  \end{minipage}\hspace{\gap}
  \begin{minipage}[t]{0.33\linewidth}\centering
    \vspace{4pt}
    \begin{minipage}[c][\teaserH][c]{\linewidth}\centering
      \animategraphics[autoplay,loop,poster=first,
        width=\linewidth,keepaspectratio]{15}{fig/ours_frames/vidB_}{0001}{0016}
    \end{minipage}
    \par\vspace{2pt}\parbox[t]{\linewidth}{\centering \textbf{(b)}~Fixed-view rendered video of the reconstructed dynamic scene}
  \end{minipage}\hspace{\gap}
  \begin{minipage}[t]{0.32\linewidth}\centering
    \vspace{0pt}
    \begin{minipage}[c][\teaserH][c]{\linewidth}\centering
      \includegraphics[width=\linewidth,height=\teaserH,keepaspectratio]{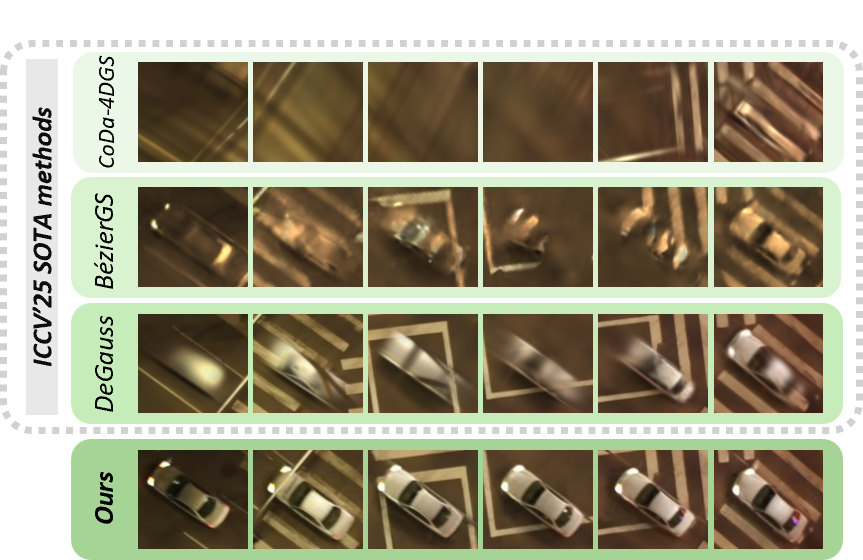}
    \end{minipage}
    \par\vspace{6pt}\parbox[t]{\linewidth}{\centering \textbf{(c)}~Vehicle trajectory comparison in the green box of (b) with SOTA}
  \end{minipage}

  \vspace{0.3em}
  \captionof{figure}{\textbf{Summary.} Given (a)~a monocular aerial video of dynamic urban scenes, AeroDGS reconstructs a physically consistent 4D model by jointly integrating static structures and dynamic motion with Gaussian representation. The framework (b)~performs photorealistic novel-view synthesis with temporally coherent geometry and (c)~achieves higher reconstruction fidelity compared to state-of-the-art methods. Please use \textbf{Adobe Reader / PDF-XChange Editor} to see animations.}
  \label{fig:teaser}
\end{center}
\endgroup

\vspace{-0.2em}
}]

\begin{abstract}
Recent advances in 4D scene reconstruction have significantly improved dynamic modeling across various domains. However, existing approaches remain limited under aerial conditions with single-view capture, wide spatial range, and dynamic objects of limited spatial footprint and large motion disparity. These challenges cause severe depth ambiguity and unstable motion estimation, making monocular aerial reconstruction inherently ill-posed.
To this end, we present \textbf{AeroDGS}, a physics-guided 4D Gaussian splatting framework for monocular UAV videos. AeroDGS introduces a Monocular Geometry Lifting module that reconstructs reliable static and dynamic geometry from a single aerial sequence, providing a robust basis for dynamic estimation. To further resolve monocular ambiguity, we propose a Physics-Guided Optimization module that incorporates differentiable ground-support, upright-stability, and trajectory-smoothness priors, transforming ambiguous image cues into physically consistent motion.
The framework jointly refines static backgrounds and dynamic entities with stable geometry and coherent temporal evolution. We additionally build a real-world UAV dataset that spans various altitudes and motion conditions to evaluate dynamic aerial reconstruction. Experiments on synthetic and real UAV scenes demonstrate that AeroDGS outperforms state-of-the-art methods, achieving superior reconstruction fidelity in dynamic aerial environments.

\vspace{-1.2em}
\end{abstract}

\section{Introduction}
\label{sec:intro}

\noindent
Aerial-based 4D reconstruction provides a unique vantage for urban perception by offering wide spatial coverage, unobstructed viewpoints and temporally continuous observations \cite{shan2023democratizing,Tang_2025_CVPR}. These advantages make it an appealing foundation for applications such as large-scale scene understanding \cite{dong2018learning}, autonomous navigation \cite{tancik2022block}, digital twin construction \cite{chaturvedi2016integrating}, and city-level dynamic perception \cite{zhou2024hugs}. However, aerial 4D reconstruction also faces severe challenges \cite{Tang_2025_CVPR} due to the intrinsic limitations of UAV imaging. Lightweight platforms typically operate with a single monocular camera, offering narrow baselines and limited parallax. The large flight altitude introduces wide depth variation, while dynamic objects appear small, move rapidly, and undergo strong illumination and appearance changes across frames \cite{maxey2024uav}. These factors together make accurate depth estimation and motion recovery in aerial monocular settings highly ill-posed.

Recent explicit Gaussian representation \cite{kerbl3Dgaussians} has opened new possibilities for efficient scene reconstruction.
By encoding geometry and appearance through spatially distributed Gaussian primitives, it enables efficient optimization and real-time rendering with fine geometric and photometric detail.
Building upon this formulation, 4DGS \cite{4dgs} extends Gaussian Splatting from static to dynamic scenes, achieving high-fidelity temporal modeling.
More advances \cite{4dgf,coda4dgs, beziergs, degauss, yan2024street} further demonstrate strong scalability and visual quality in large-scale outdoor environments.
In parallel, several feed-forward based methods \cite{wang2025vggt,dust3r_cvpr24,yao2025uni4d} are proposed for fast 3D and 4D reconstruction from single-view videos, showing strong potential for efficient scene recovery.

However, these approaches remain limited under aerial conditions.
Indoor methods \cite{4dgs, yang2023deformable3dgs} are typically designed for small-scale scenes with large articulated motions and controlled illumination, and thus always struggle to generalize to large-scale outdoor environments with varying lighting and small fast-moving objects of limited spatial footprint.
Ground-based outdoor frameworks \cite{4dgf,beziergs} rely on multi-view and LiDAR supervision, which is often impractical for lightweight UAV platforms operating at high altitude.
Due to the lack of large-scale aerial datasets, most feed-forward monocular models are trained on ground-level scenes \cite{wang2025vggt,yao2025uni4d}, leading to unreliable motion recovery and inaccurate dynamic localization when applied to aerial imagery.
As a result, monocular aerial 4D reconstruction remains an open and fundamental challenge.

To this end, we propose AeroDGS, a physics-guided 4D Gaussian Splatting framework designed for monocular UAV videos. The central idea is that while depth cues in aerial imagery are inherently ambiguous, urban scenes are not arbitrary. Since their geometry is shaped by persistent physical and structural regularities, buildings, roads, and dynamic objects follow stable spatial relations that define a low-dimensional manifold of plausible configurations. AeroDGS proposes a Physics-Guided Optimization module that encodes these regularities as differentiable constraints, guiding each object to maintain ground support, upright stability, and trajectory smoothness. These physics-guided objectives are coupled with a Monocular Geometry Lifting module that refines both static geometry and dynamic structure from raw monocular cues. Together, they transform inherently ambiguous monocular observations into consistent 4D reconstructions with stable geometry and coherent motion. We summarize our main contributions as follows:
\begin{itemize}
    \item We present \textbf{AeroDGS}, a novel 4D Gaussian Splatting framework for single-sequence aerial dynamic reconstruction. 
    The framework integrates object decomposition, temporal association, and Gaussian-based optimization to recover both static background and dynamic motion in urban environments.
    
    \item We propose a Physics-Guided Optimization module as a monocular regularization paradigm. It embeds physics-guided priors into differentiable optimization for aerial monocular reconstruction, enabling stable recovery of moving objects under single-view ambiguity.
    
    \item We construct a real-world Aero4D dataset that captures representative urban layouts and motion patterns, offering rich geometric and semantic context. The dataset serves as a benchmark for advancing aerial 4D reconstruction under realistic monocular conditions.
\end{itemize}

\noindent We demonstrate that AeroDGS delivers high-quality reconstruction of dynamic urban scenes and outperforms existing methods across both synthetic and real UAV scenarios, achieving state-of-the-art performance in dynamic aerial 4D reconstruction.

\begin{figure*}[t]
  \centering
  \includegraphics[width=\linewidth]{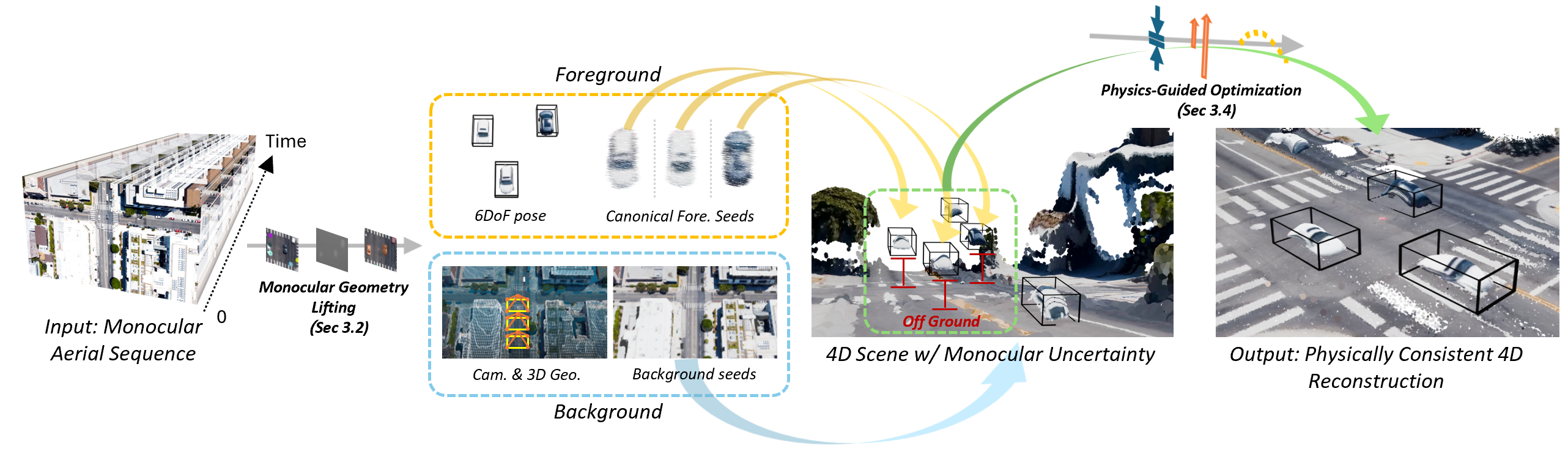}
  \caption{
\textbf{Overview of the proposed AeroDGS.} Given a monocular aerial sequence, AeroDGS introduces a \textbf{Monocular Geometry Lifting} module to reconstruct scene geometry and separate dynamic foreground from static background. The recovered seeds are composed and jointly optimized in a unified Gaussian representation. A \textbf{Physics-Guided Optimization} module is proposed to resolve pose ambiguity of dynamic objects under monocular settings, ensuring physically consistent 4D reconstruction.}
   \label{fig:overview}
\end{figure*}

\section{Related Work}
\label{sec:rw}

\subsection{Urban Scene Reconstruction}
\noindent
Urban reconstruction has been extensively studied for decades \cite{zhao2023review}. Early pipelines such as Structure-from-Motion \cite{schoenberger2016sfm} and Multi-View Stereo \cite{schoenberger2016mvs} enable large-area geometry recovery from multi-view imagery. With NeRF \cite{mildenhall2020nerf} and its urban variants \cite{martin2021nerf,turki2022mega}, the task is reformulated as volumetric neural rendering, which improves realism and robustness to outdoor appearance changes.
To enhance efficiency and scalability, 3DGS \cite{kerbl3Dgaussians} introduces explicit neural primitives. It replaces dense volumetric sampling with spatially distributed Gaussians, achieving real-time rendering and compact storage. Later frameworks \cite{4dgf,coda4dgs, beziergs, degauss, yan2024street} further explore hierarchical optimization and temporal consistency, showing that explicit primitives can achieve a good balance between quality, speed, and extensibility at the city scale. Meanwhile, feed-forward neural models \cite{wang2025vggt,dust3r_cvpr24,yao2025uni4d} have emerged that learn to predict geometric and appearance information in a single pass, enabling fast reconstruction from sparse observations.

These advances are applied in both ground and aerial settings. Ground-based systems for autonomous driving benefit from dense coverage and stable trajectories, while aerial reconstruction from UAV and satellite imagery provides broad spatial reach and unobstructed viewpoints. 
Despite extensive research, most aerial reconstructions \cite{turki2022mega,Lin_2024_CVPR, liu2024citygaussianv2, maxey2024uav} still focus on static mapping, recovering buildings, terrain, and infrastructure while filtering out moving elements. The limited parallax and small footprint of dynamic objects make consistent temporal modeling especially challenging. 
A recent work \cite{choi2025uav4d} explores pedestrian reconstruction from low-altitude UAV videos and demonstrates promising performance. 
It relies on mesh-based surface modeling tailored to close-range human motion capture. 
Building on these foundations, we present a novel framework for dynamic aerial reconstruction of urban scenes, which jointly represents static structures and temporally coherent motion.

\subsection{Dynamic Scene Representations}
\noindent
Modeling time-varying scenes is a key challenge in 3D reconstruction. Early NeRF-based work \cite{li2021neural, pumarola2021d,tretschk2021non} extended the static formulation by introducing canonical spaces and deformation fields. These approaches achieve high-quality dynamic view synthesis but are computationally expensive and difficult to scale.
Subsequent studies \cite{cao2023hexplane,fridovich2023k} improve efficiency through structured encodings.  More recently, explicit neural primitives are adopted for dynamic modeling. 4DGS \cite{4dgs} represents motion using Gaussian deformation fields, while frameworks integrate motion regularization \cite{degauss}, parametric trajectories \cite{4dgf,beziergs}, and contextual reasoning \cite{coda4dgs} to achieve real-time rendering and temporally consistent reconstruction. Complementary to these optimization-based pipelines, feed-forward neural methods \cite{wang2025vggt, yao2025uni4d} leverage learned scene priors to estimate dynamic geometry and motion efficiently, bypassing iterative optimization across time.

Despite this progress, most optimization-based dynamic representations \cite{4dgs, 4dgf} rely on multi-view inputs to provide geometric priors for motion estimation. Such priors help disambiguate object trajectories and maintain spatial consistency, but they are difficult to obtain in aerial imagery captured by a single monocular camera. 
Meanwhile, several feed-forward methods have been proposed for dynamic reconstruction, yet most \cite{wang2025vggt,dust3r_cvpr24} are trained on ground-based data and fail to generalize to aerial settings due to limited training resources.
In UAV scenes, dynamic objects often occupy only a small fraction of the image, and their locations along depth are inherently ambiguous, making it challenging to recover accurate and consistent motion. To this end, we design a set of physically guided regularization terms to resolve the positional ambiguity of dynamic objects in monocular settings, reducing their degrees of freedom to conform to real-world physical rules and improving reconstruction stability.

\subsection{UAV Dataset}
\noindent
Progress in aerial vision research is driven by datasets. Most existing benchmarks \cite{cao2021visdrone, du2018unmanned, barekatain2017okutama} provide large-scale annotated imagery for 2D perception tasks.
Several subsequent datasets \cite{chen2022stpls3d, wang2025uavscenes} extend this line of research toward 3D geometry recovery and static reconstruction, providing dense point clouds and multi-view imagery to benchmark aerial mapping, surface reconstruction, and localization tasks.
However, datasets for aerial 4D reconstruction remain scarce. Partial simulation-based resources \cite{uav3d2024,hu2023aerial} offer synthetic aerial scenes that can approximately support spatio-temporal modeling, yet they are generated in virtual environments and contain discrete image sequences where adjacent frames are captured from widely separated viewpoints. The large spatial gaps and lack of natural motion continuity prevent them from representing realistic aerial dynamics. Moreover, collecting real dynamic UAV data is also challenging due to flight regulations, costly annotation, and the difficulty of maintaining precise temporal alignment and object consistency \cite{rizzoli2023syndrone}.
To address this gap, we construct a representative UAV dataset comprising multiple aerial sequences with both static and dynamic components, serving as a benchmark for dynamic aerial reconstruction under realistic flight conditions.

\begin{figure}[t]
  \centering
  \includegraphics[width=\linewidth]{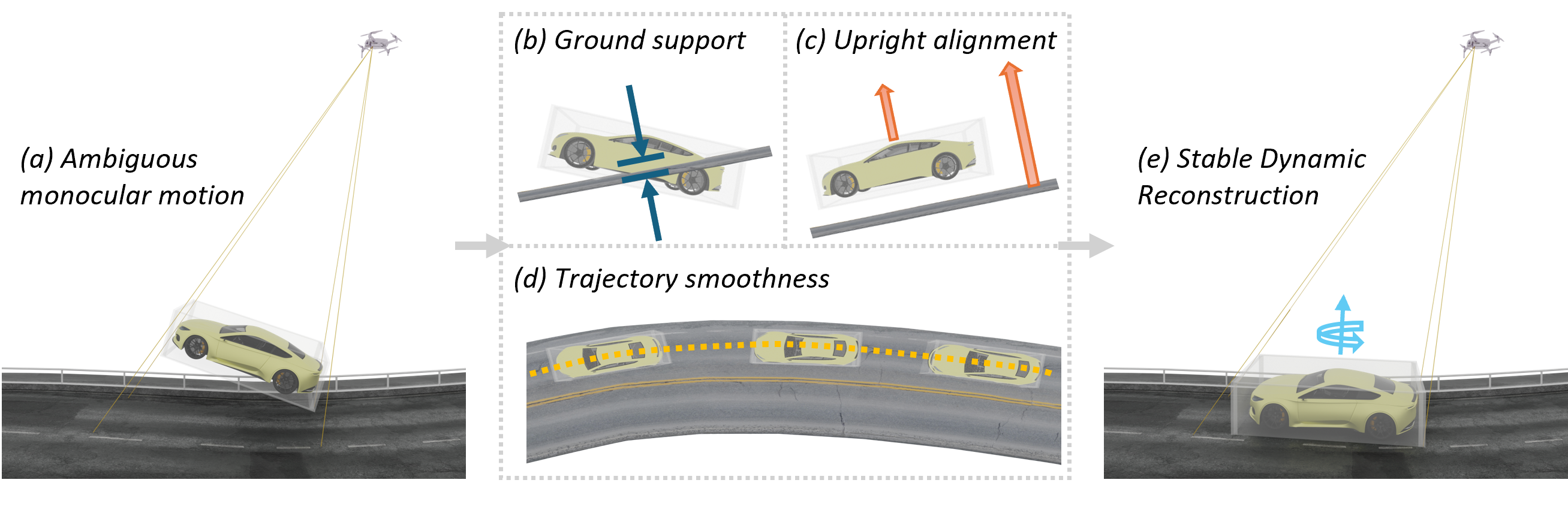}
  \caption{
\textbf{Physics-Guided Optimization.}
(a) In monocular UAV scenes, dynamic objects exhibit uncertain 3D positions and orientations due to single-view geometry and small image footprints. 
AeroDGS introduces differentiable physics-guided constraints that enforce 
(b) \textbf{ground support}, maintaining consistent contact with the local plane; 
(c) \textbf{upright stability}, aligning the vertical axis with the reference direction; and 
(d) \textbf{trajectory smoothness}, ensuring continuous acceleration and temporally coherent motion. 
(e) These constraints transform under-determined poses into a single real-world-consistent configuration, yielding accurate motion recovery and stable optimization.
}
  \label{fig:optimization}
\end{figure}

\section{Method}
\label{sec:method}
We present AeroDGS, a physics-guided dynamic Gaussian framework for monocular UAV videos that reconstructs physically consistent 4D scenes under real-world constraints.
An overview of our method is shown in \cref{fig:overview}.
\cref{sec:3.1} defines the problem, \cref{sec:3.2} introduces the Monocular Geometry Lifting module, \cref{sec:3.3} details the Gaussian representation, and \cref{sec:3.4} describes the proposed Physics-Guided Optimization module. 

\subsection{Problem Setup}
\label{sec:3.1}

Given a monocular UAV video sequence $\{I_t\}_{t=1}^{T}$ captured by a moving aerial camera flying over an urban area, each frame $I_t$ observes a scene composed of a static background and a set of dynamic objects $\mathcal{O}$.  
The camera intrinsics and poses $P_t \in SE(3)$ are unknown and will be estimated.  
For each dynamic object $o \in \mathcal{O}$, its 3D center $c_{o,t}$ and transformation $T_{o,t} \in SE(3)$ describe its motion through time.  
The goal is to reconstruct a temporally coherent 4D scene representation that enables photorealistic novel-view rendering and consistent recovery of camera trajectories, scene geometry, and dynamic motion from monocular input.

\subsection{Monocular Geometry Lifting}
\label{sec:3.2}

Conventional Structure-from-Motion pipelines \cite{schoenberger2016sfm,schoenberger2016mvs} often fail to reconstruct dynamic objects and produce only sparse ground points, lacking the geometric priors required for subsequent motion reasoning and physical constraints \cite{bescos2018dynaslam}.
Therefore, we introduce a Monocular Geometry Lifting module that recovers complete 3D geometry, camera poses, and dynamic instances from a single sequence input.  

We first leverage zero-shot 2D foundation cues to obtain coarse semantic and structural priors for subsequent reconstruction.
For each frame $I_t$, a depth estimator \cite{piccinelli2024unidepth} provides a dense pseudo-depth map $D_t$ and camera intrinsics $K$.  
Segmentation \cite{ravi2024sam2, liu2023grounding} and tracking \cite{cheng2023tracking} of potentially movable instances produce 2D correspondences across all frames, maintaining consistent object identities over time.  
Meanwhile, long-term background feature tracks $\{k_{t,j}\}$ \cite{karaev2024cotracker3} are triangulated and refined through local bundle adjustment \cite{agarwal2010bundle} to estimate scale-consistent keypoints and camera poses $P_t = [R_t \,|\, t_t]$.  

To correct scale variations in monocular depth, a local ratio field between geometric and predicted depths is computed on tracked points and interpolated to yield a refined depth $\tilde{D}_t$.  
Each pixel $x$ is then back-projected into 3D as  
\begin{equation}
X_t(x) = \Pi^{-1}(x, \tilde{D}_t(x), K),
\end{equation}
forming a point map that jointly represents static and potentially dynamic regions.  
Pixels belonging to the same 2D instance are grouped into an object-level point set $P_{o,t}$.  
For each object, an oriented bounding box is fitted via PCA \cite{abdi2010principal} to estimate its 3D center $c_{o,t}$ and footprint size $(w,\ell)$. The height $h$ is predicted by a pretrained MLP, since each moving object is captured under single-view geometry in the monocular setting, where depth cannot be inferred due to the absence of multi-view constraints.
ID switches and occlusions in 2D instance tracking are resolved in 3D space by grounding objects on physically plausible positions along camera rays and enforcing trajectory smoothness.
Objects with 3D displacements below a threshold are regarded as static to reduce unnecessary tracking overhead, and the rest are initialized as dynamic candidates.  
This process provides dense ground geometry and temporally consistent object priors as a robust foundation for subsequent optimization.

\subsection{Scene Representation}
\label{sec:3.3}
We represent both static backgrounds and dynamic foregrounds using explicit 3D Gaussian primitives \cite{kerbl3Dgaussians}. This unified and differentiable representation enables object-level motion optimization while preserving geometric consistency with the static scene. 

\paragraph{Gaussian primitives.}  
Each primitive is defined by its center position $\mu_i \in \mathbb{R}^3$, covariance $\Sigma_i \in \mathbb{R}^{3\times3}$, opacity $\alpha_i$, and appearance $A_i$. Aerial imagery exhibits large appearance variation caused by changing sunlight direction, surface reflection, and camera motion. Inspired by \cite{4dgf}, we model the appearance of each Gaussian as a continuous field:
\begin{equation}
A_i = f_\phi(\mu_i, d, t, e_o),
\end{equation}
where $f_\phi$ denotes a shared appearance field conditioned on spatial position $\mu_i$, viewing direction $d$, and temporal index $t$, with an additional embedding $e_o$ for dynamic instances.  
The field is parameterized through three complementary encodings. Spatial coordinates are normalized within the scene bounds and embedded through a hash grid \cite{mueller2022instant} to capture localized spatial variation. Directional dependency is represented by a spherical-harmonic basis to account for view-related appearance changes, while temporal information is encoded by sinusoidal embeddings that maintain temporal ordering and support smooth appearance evolution.  
This joint parameterization models the rich appearance variations of urban environments without storing per-Gaussian spherical harmonics, enhancing temporal coherence and reducing memory cost for denser Gaussian representation.

\paragraph{Dynamic object encoding.}  

For each moving instance $o$, a subset of Gaussians $\mathcal{G}_o$ is defined in its canonical object space bounded by an axis-aligned box with dimensions $(w, \ell, h)$. 
The motion of each object is modeled as a continuous 6DoF trajectory in the Lie group $SE(3)$,
\begin{equation}
T_{o,t} = \exp\!\big(\xi_o(t)\big),
\end{equation}
where $\xi_o(t) \in \mathfrak{se}(3)$ denotes the time-dependent twist vector describing rotational and translational components. 
To account for small deviations and pose uncertainties, a residual correction $\Delta T_{o,t}$ is introduced and optimized jointly with other parameters,
\begin{equation}
T'_{o,t} = \Delta T_{o,t} \cdot T_{o,t}.
\end{equation}
Each Gaussian center is then transformed into the world coordinate system as
\begin{equation}
\mu_{i,t} = T'_{o,t} \circ \mu_i.
\end{equation}
Temporal continuity between timestamps is achieved by spline interpolation in $SE(3)$, ensuring smooth and physically consistent trajectories.  

\paragraph{Scene composition.}  
At each timestamp $t$, the scene is represented as the union of a static Gaussian set and dynamic subsets. The static component $\mathcal{G}_{\text{static}}$ encodes background geometry and appearance, while each dynamic subset $\mathcal{G}_o$ evolves over time through its transformation $T'_{o,t}$. 
The overall scene at time $t$ is formulated as  
\begin{equation}
\mathcal{G}(t)=\mathcal{G}_{\text{static}} \cup 
\bigcup_{o\in\mathcal{O}}T'_{o,t}\circ\mathcal{G}_o.
\end{equation}
To ensure physical consistency under monocular reconstruction, we introduce a Physics-Guided Optimization module that reduces motion uncertainty of dynamic objects based on real-world physical constraints. 
A local ground plane $\Gamma_t$ is inferred from static geometry to guide dynamic motion, ensuring physically plausible trajectories while allowing local flexibility.
The detailed formulations are described in \cref{sec:3.4} and illustrated in \cref{fig:optimization}.

\paragraph{Rendering.}
The composed scene $\mathcal{G}(t)$ is rendered following the differentiable rasterization of 3DGS \cite{kerbl3Dgaussians}. 
Each primitive projects to a 2D Gaussian on the image plane, and pixel colors are obtained by standard $\alpha$-compositing:
\begin{equation}
I_t(x)=\sum_{i}\Bigg[\;\alpha_i(x)\,A_i(x)\;\prod_{j<i}\big(1-\alpha_j(x)\big)\Bigg].
\end{equation}

\subsection{Physics-Guided Optimization}
\label{sec:3.4}

\paragraph{Initialization.}
The optimization begins with the result of the Monocular Geometry Lifting.
Static Gaussians are defined in the world coordinate system, where each reconstructed 3D point provides the mean $\boldsymbol{\mu}_i$, and its local spatial distribution determines the covariance $\boldsymbol{\Sigma}_i$. 
Dynamic Gaussians are represented in canonical object spaces and initialized from the 3D points within the estimated object bounding regions. 
Camera poses $P_t$ obtained from bundle adjustment are refined by a learnable residual $\Delta P_t$ to correct small frame misalignments. 

\paragraph{Overall objective.}
We couple photometric supervision with physics-guided regularization to jointly optimize appearance fidelity and physically consistent motion. The total loss is defined as:
\begin{equation}
\mathcal{L} =
\lambda_{\text{photo}}\mathcal{L}_{\text{photo}} +
\lambda_{\text{sup}}\mathcal{L}_{\text{support}} +
\lambda_{\text{upr}}\mathcal{L}_{\text{upright}} +
\lambda_{\text{traj}}\mathcal{L}_{\text{traj}} .
\end{equation}

\paragraph{Photometric supervision.}
The photometric term $\mathcal{L}_{\text{photo}}$ measures the color consistency between the rendered image $\hat{I}_t$ and the input frame $I_t$ using a combination of $L_1$ norm and SSIM losses \cite{wang2004image}. 
During the warm-up stage, static and dynamic regions are equally weighted to stabilize the optimization. 
After the static background converges, the weight of dynamic regions is increased to refine fine-scale appearance and motion.

\paragraph{Physics-guided regularization.}
To achieve geometrically and temporally consistent motion under monocular UAV input, three regularization terms are proposed.

\emph{Support consistency.}
Each dynamic object is encouraged to stay close to the estimated local ground level. 
The support loss is defined as:
\begin{equation}
\mathcal{L}_{\text{support}}
= \mathbb{E}_{o,t}\!
\left[
\psi\!\big(
\mathbf{r}_{o,t}^{\top}
(\mathbf{c}_{o,t} - \hat{\mathbf{c}}^{\,g}_{o,t})
\big)
\right],
\end{equation}
where $\psi(\cdot)$ is a robust penalty that measures the signed distance between the object center and its ground projection along the camera viewing ray $\mathbf{r}_{o,t}$. 
$\mathbf{c}_{o,t}$ denotes the 3D center of object $o$ at time $t$, and $\hat{\mathbf{c}}^{\,g}_{o,t}$ is its intersection with the local ground plane. 
In our formulation, the constraint is applied to $\mathbf{c}_{o,t}$ shifted upward by half of the object height, ensuring that the bottom of the vehicle stays sufficiently close to the ground along the ray direction while tolerating small reconstruction noise.

\emph{Upright stability.}
The upright constraint aligns the vertical axis of each object with the reference direction:
\begin{equation}
\mathcal{L}_{\text{upright}}
= \mathbb{E}_{o,t}\!
\left[
1 - |\mathbf{u}_{o,t}\!\cdot\!\mathbf{v}_{o,t}|
\right],
\end{equation}
where $\mathbf{u}_{o,t}$ denotes the object’s vertical axis and $\mathbf{v}_{o,t}$ is the reference vector,
defined as the ground normal $\mathbf{n}_t$ for rigid objects and the gravity direction $\mathbf{g}$ for non-rigid ones.
This loss constrains 3-DoF rotations to align with physically plausible motion by encouraging rotation around the vertical axis, reducing unrealistic tilting and facilitating stable pose optimization.

\emph{Trajectory smoothness.}
To promote temporally coherent motion, we design a second-order smoothness constraint:
\begin{equation}
\mathcal{L}_{\text{traj}}
= \mathbb{E}_{o,t}\!
\left[
\|\mathbf{c}_{o,t+1} - 2\mathbf{c}_{o,t} + \mathbf{c}_{o,t-1}\|_2^2
\right].
\end{equation}
This loss suppresses high-frequency jitter and enforces continuous acceleration.
It also allows objects exiting the scene to retain their motion momentum, ensuring they move naturally out of view rather than abruptly stopping at the scene boundary.

\section{Dataset}
\label{sec:dataset}

Real-world monocular UAV data for dynamic 4D reconstruction remain scarce. Existing related aerial datasets \cite{uav3d2024,rizzoli2023syndrone} are synthetic and lack temporally continuous viewpoints, limiting their suitability for real-world dynamic modeling. To address this gap, we build the Aero4D dataset, a curated and compact validation dataset consisting of UAV sequences that capture realistic flight trajectories, temporal continuity, and varied dynamic scenes.

\vspace{-0.5cm}
\paragraph{Data collection.}
The dataset combines aerial data collected in urban environments with representative aerial footage selected from online sources. All sequences are recorded at resolutions from 2K to 4K, at 15\,fps, with flight altitudes between 50\,m and 100\,m, covering areas of approximately 2,000--18,000\,m$^2$ and a ground sampling distance of 3--5\,cm/px. The scenes include varied illumination and motion conditions, with the aim of reflecting typical UAV observation settings.

\vspace{-0.5cm}
\paragraph{Annotations.}
Each frame is associated with a camera pose and a dense point cloud representing the static environment. Dynamic instances are annotated with pixel-level instance masks and temporally consistent IDs. For each instance, we provide 3D bounding boxes with 6-DoF poses and trajectories defined in a global coordinate system. Please refer to the supplementary material for more details.

\section{Experiments}
\label{sec:exp}

\begin{table*}[t]
\centering
\footnotesize
\setlength{\tabcolsep}{2.8pt}
\renewcommand{\arraystretch}{1.12}
\caption{
\textbf{Novel-view synthesis results on Aero4D dataset.} 
\textbf{AeroDGS} performs better than state-of-the-art methods under varying altitudes, illumination, and real-world conditions. }
\label{tab:aero4d}
\resizebox{\linewidth}{!}{
\begin{tabular}{lcccccccccccc}
\toprule
 & \multicolumn{4}{c}{Intersection-Night} & \multicolumn{4}{c}{Downtown-High} & \multicolumn{4}{c}{Intersection-Day} \\
\cmidrule(lr){2-5} \cmidrule(lr){6-9} \cmidrule(lr){10-13}
Method &
PSNR$\uparrow$ & SSIM$\uparrow$ & LPIPS$\downarrow$ & Dyn-PSNR$\uparrow$ &
PSNR$\uparrow$ & SSIM$\uparrow$ & LPIPS$\downarrow$ & Dyn-PSNR$\uparrow$ &
PSNR$\uparrow$ & SSIM$\uparrow$ & LPIPS$\downarrow$ & Dyn-PSNR$\uparrow$ \\
\midrule
4DGS \cite{4dgs}     & 22.02 & 0.808 & 0.420 & 11.74 & 23.06 & 0.784 & 0.402 & 10.50 & 19.33 & 0.729 & 0.644 & - \\
BézierGS \cite{beziergs} & 21.89 & 0.794 & 0.295 & \third{12.50} & 19.51 & 0.726 & 0.366 & - & \third{28.31} & \third{0.928} & \third{0.120} & - \\
CoDa-4DGS \cite{coda4dgs} & 23.10 & 0.834 & 0.335 & - & 15.37 & 0.583 & 0.652 & 9.56 & 19.75 & 0.727 & 0.521 & - \\
DeGauss \cite{degauss} & \third{27.90} & \third{0.925} & \third{0.114} & - & \third{28.31} & \third{0.908} & \third{0.110} & \second{18.70} & 28.00 & 0.880 & 0.194 & \third{15.30} \\
4DGF  \cite{4dgf}  & \second{30.54} & \second{0.969} & \second{0.036} & \second{13.48} & \second{35.38} & \best{0.977} & \second{0.014} & \third{17.94} & \second{32.35} & \second{0.967} & \second{0.023} & \second{17.26} \\
\textbf{AeroDGS (Ours)} &
\best{32.71} & \best{0.971} & \best{0.024} & \best{17.65} &
\best{37.91} & \second{0.974} & \best{0.013} & \best{22.47} &
\best{34.84} & \best{0.971} & \best{0.018} & \best{21.75} \\
\bottomrule
\end{tabular}}
\end{table*}

\begin{table}[t]
\centering
\footnotesize
\setlength{\tabcolsep}{3.5pt}
\renewcommand{\arraystretch}{1.1}
\caption{
\textbf{Novel-view synthesis results on the challenging synthetic UAV3D \cite{uav3d2024} dataset.} 
\textbf{AeroDGS} outperforms the state of the art, with notably better performance in dynamic scene regions. 
}
\label{tab:uav3d}
\begin{tabular}{lcccc}
\toprule
Method & PSNR$\uparrow$ & SSIM$\uparrow$ & LPIPS$\downarrow$ & Dyn-PSNR$\uparrow$ \\
\midrule
4DGS  \cite{4dgs}    & 11.65 & 0.326 & 0.742 & - \\
BézierGS \cite{beziergs}  & 10.34 & 0.303 & 0.620 & \third{7.16} \\
CoDa-4DGS \cite{coda4dgs} & 13.36 & 0.385 & 0.602 & - \\
DeGauss  \cite{degauss} & \third{23.53} & \third{0.834} & \third{0.139} & - \\
4DGF   \cite{4dgf}   & \second{31.78} & \second{0.968} & \second{0.031} & \second{10.63} \\
\textbf{AeroDGS (Ours)} & \best{33.61} & \best{0.972} & \best{0.026} & \best{15.60} \\
\bottomrule
\end{tabular}
\end{table}

\paragraph{Scenes.} 
We evaluate our method on both synthetic and real UAV benchmarks. 
For the synthetic dataset UAV3D \cite{uav3d2024}, we use the sequences in Town03, which contain a relatively large number of dynamic objects and diverse motion patterns. 
For real-world evaluation, we select three representative UAV scenes: 
Intersection-Night, a low-altitude nighttime intersection with dense traffic and specular reflections; 
Downtown-High, a high-altitude flight over urban blocks where dynamic objects appear small; 
and Intersection-Day, a daytime intersection featuring vehicles that slowly start or stop near traffic lights.
These scenes together cover diverse altitudes, illumination conditions, and traffic dynamics for comprehensive evaluation.

\vspace{-0.4cm}
\paragraph{Evaluation metrics.} 
Following prior work, we evaluate novel-view image quality using PSNR, SSIM \cite{wang2004image}, and LPIPS \cite{zhang2018unreasonable}, 
and assess dynamic fidelity with Dyn-PSNR, computed as the PSNR within dynamic regions.
All reported results are averaged over three independent runs for consistency.

\subsection{Implementation Details}
Experiments are conducted on a single NVIDIA RTX~6000~Ada GPU with 48\,GB of memory. 
Training is performed for 30K iterations using the Adam optimizer \cite{kingma2017adammethodstochasticoptimization}. 
The loss weights are set to $\lambda_{\text{photo}}=1.0$, $\lambda_{\text{sup}}=0.05$, 
$\lambda_{\text{upr}}=0.1$, and $\lambda_{\text{traj}}=0.02$. 
The dynamic-mask weighting increases linearly from 1.0 to 1.3 between 7K and 15K iterations. 
Meanwhile, the physics-guided constraints decay by 50\% over the same period. 
Training uses full-resolution images, with an 8:2 split between training and validation frames. 

\subsection{Comparison to State-of-the-Art}
We compare our method with recent state-of-the-art approaches in dynamic 4D reconstruction, 
including 
BézierGS \cite{beziergs}, 
CoDa-4DGS \cite{coda4dgs},
DeGauss \cite{degauss},
4DGF \cite{4dgf},
and 4DGS \cite{4dgs}.
Our method operates on monocular video input and internally estimates scene geometry, whereas baselines require external priors. 
Accordingly, baselines are evaluated using the priors provided with each dataset following their standard protocols. 
For the Aero4D dataset, although these priors are carefully refined during dataset construction, minor residual inaccuracies may still persist due to the inherent ambiguity of monocular aerial capture, consistent with realistic UAV conditions.

\vspace{-0.3cm}
\paragraph{Quantitative results.} \cref{tab:aero4d} and \cref{tab:uav3d} summarize the novel-view synthesis results on the synthetic and real-world UAV datasets, respectively.
Our method achieves the best overall performance across the metrics, 
outperforming previous approaches by up to 4\,dB on dynamic regions. 

Although our method achieves the best dynamic quality among all baselines, 
we observe that the gap between static and dynamic metrics is relatively large, 
which should not be interpreted as a direct quality difference. 
Due to the nonlinear motion of the UAV platform and moving objects, 
the estimated positions of dynamic objects in the novel views may exhibit slight deviations from the ground truth locations. 
For small moving objects that occupy only a limited number of pixels in the image, such positional differences can lead to a noticeable drop in PSNR, even when the perceptual quality remains comparable. This effect is further reflected in our qualitative results, which provide additional evidence of the robustness of our dynamic reconstruction. 

Notably, the overall metrics of the synthetic benchmark are lower than those of the real-world sequences. 
This results from discrete camera trajectories and long-baseline viewpoint shifts in the synthetic UAV3D \cite{uav3d2024} dataset, 
which amplify temporal misalignment and reveal the lack of a dedicated, high-quality dataset for aerial 4D reconstruction. 

\begin{figure*}[t]
  \centering
  \includegraphics[width=\linewidth]{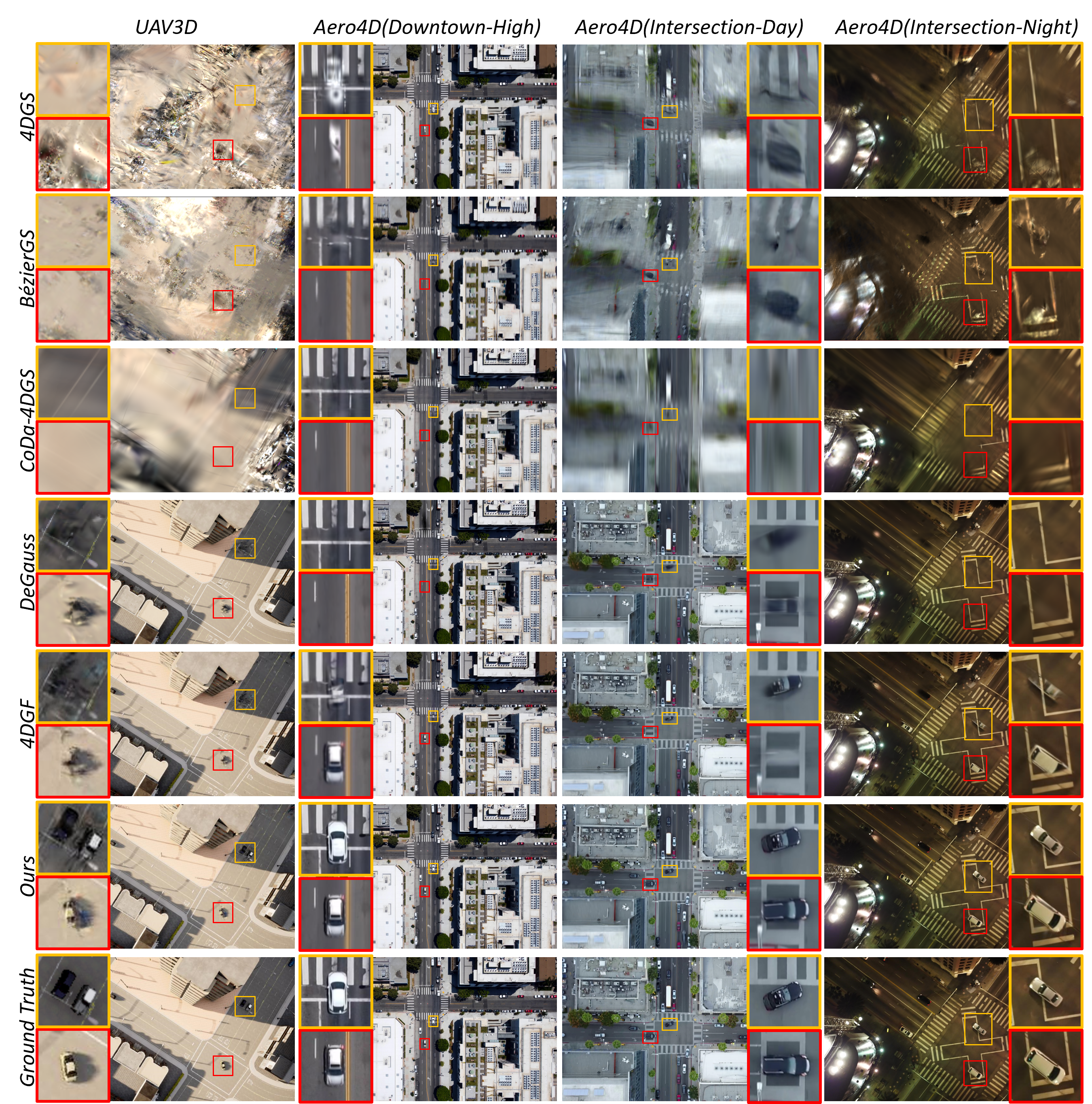}
  \caption{
  \textbf{Qualitative comparison of novel-view synthesis results.} 
Our method achieves high overall reconstruction quality on both synthetic and real-world UAV datasets, maintaining high fidelity under diverse altitudes, illumination, and object motion patterns. 
Sharper structures and more consistent appearance are preserved compared with state-of-the-art methods. 
Yellow and red rectangular boxes highlight enlarged views of corresponding areas for visual comparison.}
  \label{fig:results}
\end{figure*}

\vspace{-0.4cm}
\paragraph{Qualitative results.}
Novel view synthesis results on the synthetic UAV3D \cite{uav3d2024} and real-world Aero4D dataset are shown in Figure~\ref{fig:results}. 
For the synthetic scenario, 4DGS \cite{4dgs}, BézierGS \cite{beziergs}, and CoDa-4DGS \cite{coda4dgs} exhibit noticeable blurring and incomplete geometry under wide-baseline viewpoints, 
while DeGauss \cite{degauss} and 4DGF \cite{4dgf} successfully reconstruct static structures but struggle to recover dynamic objects. 
In contrast, our method produces more complete and consistent reconstructions for both static and dynamic components. 
For the real-world scenes,
our method reconstructs dynamic objects with sharper details and clearer shapes than previous art across diverse altitudes, illumination conditions, and motion ranges, demonstrating the effectiveness of our dynamic object representation. 

\begin{table}[t]
\centering
\caption{\textbf{Ablation study.} Novel view synthesis on the Aero4D dataset.}
\label{tab:ablation}
\resizebox{0.95\linewidth}{!}{%
\begin{tabular}{lcccc}
\toprule
 & PSNR$\uparrow$ & SSIM$\uparrow$ & LPIPS$\downarrow$ & Dyn-PSNR$\uparrow$ \\
\midrule
w/o Initialization    & 33.82 & 0.952 & 0.027 & 17.75 \\
w/o Ground Support    & \third{34.51} & \second{0.967} & \second{0.022} & 19.23 \\
w/o Upright Stability & 34.10 & 0.963 & 0.025 & \third{19.35} \\
w/o Trajectory Smoothness & \second{34.63} & \third{0.966} & \second{0.022} & \second{19.89} \\
w/o Dynamic Mask      & 34.23 & 0.959 & 0.024 & 19.12 \\
Ours (Full Model)     & \best{34.67} & \best{0.971} & \best{0.021} & \best{20.07} \\
\bottomrule
\end{tabular}}
\end{table}

\subsection{Ablation Studies}
\paragraph{Scene initialization.}
We compare our Monocular Geometry Lifting module with an SfM-based setup.
In the baseline, sparse points and camera poses are reconstructed by COLMAP \cite{schoenberger2016sfm}, and a ground plane is fitted from the sparse points.
Dynamic objects are localized by projecting their 2D mask centers onto this plane along camera rays, estimating 3D position, moving direction, and size from the back-projected region with a fixed height prior.
Tracking is obtained by associating 3D centers across frames.

This initialization provides only a coarse prior, which introduces plane bias and consequently leads to pose inaccuracies for dynamic objects.
As shown in \cref{tab:ablation}, the dynamic PSNR drops noticeably.

\vspace{-0.3cm}
\paragraph{Physics-Guided optimization.}
\cref{tab:ablation} further demonstrates the contribution of each physics-guided constraint and the dynamic-mask weighting.
Removing the ground-support, upright-stability, or trajectory-smoothness term degrades reconstruction fidelity and temporal coherence, confirming that each physically grounded constraint stabilizes monocular optimization and enforces realistic motion.
In addition, removing the dynamic-mask weighting weakens small-object reconstruction, as it reduces attention to fine-scale motion in aerial views.
The full model achieves the best overall performance across the metrics, validating the effectiveness of the proposed regularization module.

\vspace{-0.1cm}
\subsection{Limitation}
Despite the strong performance of AeroDGS, several limitations remain.
First, our current 3\,\text{m}
 motion-threshold strategy for dynamic–static separation may misclassify objects with small localized motion, causing them to be processed by the static pipeline and appear blurred in the rendered results.
Second, we do not reconstruct pedestrians, as they appear only partially with extremely limited pixel coverage in high-altitude aerial views.
\vspace{-0.1cm}

\vspace{-2cm}
\enlargethispage{1\baselineskip}

\section{Conclusion}
\label{sec:con}
\vspace{-0.3cm}
We present AeroDGS, a physics-guided dynamic Gaussian splatting framework for monocular aerial 4D reconstruction.
The Monocular Geometry Lifting module initializes the scene geometry, while Gaussian-based optimization jointly refines static and dynamic components.
The Physics-Guided Optimization module resolves pose ambiguity of dynamic objects in monocular aerial videos through physically grounded motion modeling.
Experiments under varying flight altitudes, illumination, and motion conditions show that AeroDGS outperforms state-of-the-art methods, delivering high-fidelity reconstruction and temporally coherent dynamics.

\section{Acknowledgements}
This work was partially supported by the United States Air Force Research Laboratory and the AFRL Regional Hub under Cooperative Agreement No. FA8750-22-2-0501, the Intelligence Advanced Research Projects Activity (IARPA) via the Department of Interior/Interior Business Center (DOI/IBC) under Grant No. 140D0423C0075, and the Office of Naval Research under Grant No. N000142312670.

{
    \small
    \bibliographystyle{ieeenat_fullname}
    \bibliography{main}

@String(CVPR= {IEEE Conf. Comput. Vis. Pattern Recog.})

@String(ICCV= {Int. Conf. Comput. Vis.})

@String(ECCV= {Eur. Conf. Comput. Vis.})

@String(CVPR  = {CVPR})

@String(ICCV  = {ICCV})

@String(ECCV  = {ECCV})

@inproceedings{schoenberger2016sfm,
    author={Sch\"{o}nberger, Johannes Lutz and Frahm, Jan-Michael},
    title={Structure-from-Motion Revisited},
    booktitle={Conference on Computer Vision and Pattern Recognition (CVPR)},
    year={2016},
}

@inproceedings{schoenberger2016mvs,
    author={Sch\"{o}nberger, Johannes Lutz and Zheng, Enliang and Pollefeys, Marc and Frahm, Jan-Michael},
    title={Pixelwise View Selection for Unstructured Multi-View Stereo},
    booktitle={European Conference on Computer Vision (ECCV)},
    year={2016},
}

@article{bescos2018dynaslam,
  title={DynaSLAM: Tracking, mapping, and inpainting in dynamic scenes},
  author={Bescos, Berta and F{\'a}cil, Jos{\'e} M and Civera, Javier and Neira, Jos{\'e}},
  journal={IEEE robotics and automation letters},
  volume={3},
  number={4},
  pages={4076--4083},
  year={2018},
  publisher={IEEE}
}

@inproceedings{piccinelli2024unidepth,
    title     = {{U}ni{D}epth: Universal Monocular Metric Depth Estimation},
    author    = {Piccinelli, Luigi and Yang, Yung-Hsu and Sakaridis, Christos and Segu, Mattia and Li, Siyuan and Van Gool, Luc and Yu, Fisher},
    booktitle = {Proceedings of the IEEE/CVF Conference on Computer Vision and Pattern Recognition (CVPR)},
    year      = {2024}
}

@article{ravi2024sam2,
  title={SAM 2: Segment Anything in Images and Videos},
  author={Ravi, Nikhila and Gabeur, Valentin and Hu, Yuan-Ting and Hu, Ronghang and Ryali, Chaitanya and Ma, Tengyu and Khedr, Haitham and R{\"a}dle, Roman and Rolland, Chloe and Gustafson, Laura and Mintun, Eric and Pan, Junting and Alwala, Kalyan Vasudev and Carion, Nicolas and Wu, Chao-Yuan and Girshick, Ross and Doll{\'a}r, Piotr and Feichtenhofer, Christoph},
  journal={arXiv preprint arXiv:2408.00714},
  url={https://arxiv.org/abs/2408.00714},
  year={2024}
}

@article{liu2023grounding,
  title={Grounding dino: Marrying dino with grounded pre-training for open-set object detection},
  author={Liu, Shilong and Zeng, Zhaoyang and Ren, Tianhe and Li, Feng and Zhang, Hao and Yang, Jie and Li, Chunyuan and Yang, Jianwei and Su, Hang and Zhu, Jun and others},
  journal={arXiv preprint arXiv:2303.05499},
  year={2023}
}

@inproceedings{cheng2023tracking,
  title={Tracking Anything with Decoupled Video Segmentation},
  author={Cheng, Ho Kei and Oh, Seoung Wug and Price, Brian and Schwing, Alexander and Lee, Joon-Young},
  booktitle={ICCV},
  year={2023}
}

@inproceedings{agarwal2010bundle,
  title={Bundle adjustment in the large},
  author={Agarwal, Sameer and Snavely, Noah and Seitz, Steven M and Szeliski, Richard},
  booktitle={European conference on computer vision},
  pages={29--42},
  year={2010},
  organization={Springer}
}

@article{abdi2010principal,
  title={Principal component analysis},
  author={Abdi, Herv{\'e} and Williams, Lynne J},
  journal={Wiley interdisciplinary reviews: computational statistics},
  volume={2},
  number={4},
  pages={433--459},
  year={2010},
  publisher={Wiley Online Library}
}

@Article{kerbl3Dgaussians,
      author       = {Kerbl, Bernhard and Kopanas, Georgios and Leimk{\"u}hler, Thomas and Drettakis, George},
      title        = {3D Gaussian Splatting for Real-Time Radiance Field Rendering},
      journal      = {ACM Transactions on Graphics},
      number       = {4},
      volume       = {42},
      month        = {July},
      year         = {2023},
      url          = {https://repo-sam.inria.fr/fungraph/3d-gaussian-splatting/}
}

@InProceedings{4dgf,
    author    = {Tobias Fischer and Jonas Kulhanek and Samuel Rota Bul{\`o} and Lorenzo Porzi and Marc Pollefeys and Peter Kontschieder},
    title     = {Dynamic 3D Gaussian Fields for Urban Areas},
    booktitle = {The Thirty-eighth Annual Conference on Neural Information Processing Systems},
    year      = {2024}
}

@article{mueller2022instant,
    author = {Thomas M\"uller and Alex Evans and Christoph Schied and Alexander Keller},
    title = {Instant Neural Graphics Primitives with a Multiresolution Hash Encoding},
    journal = {ACM Trans. Graph.},
    issue_date = {July 2022},
    volume = {41},
    number = {4},
    month = jul,
    year = {2022},
    pages = {102:1--102:15},
    articleno = {102},
    numpages = {15},
    url = {https://doi.org/10.1145/3528223.3530127},
    doi = {10.1145/3528223.3530127},
    publisher = {ACM},
    address = {New York, NY, USA}
}

@article{wang2004image,
  title={Image quality assessment: from error visibility to structural similarity},
  author={Wang, Zhou and Bovik, Alan C and Sheikh, Hamid R and Simoncelli, Eero P},
  journal={IEEE transactions on image processing},
  volume={13},
  number={4},
  pages={600--612},
  year={2004},
  publisher={IEEE}
}

@inproceedings{uav3d2024,
    title={UAV3D: A Large-scale 3D Perception Benchmark for Unmanned Aerial Vehicles},
    author={Hui Ye and Raj Sunderraman and Shihao Ji},
    booktitle={The 38th Conference on Neural Information Processing Systems (NeurIPS)},
    year={2024}
}

@article{hu2023aerial,
  title={Aerial monocular 3d object detection},
  author={Hu, Yue and Fang, Shaoheng and Xie, Weidi and Chen, Siheng},
  journal={IEEE Robotics and Automation Letters},
  volume={8},
  number={4},
  pages={1959--1966},
  year={2023},
  publisher={IEEE}
}

@inproceedings{rizzoli2023syndrone,
  title={Syndrone-multi-modal uav dataset for urban scenarios},
  author={Rizzoli, Giulia and Barbato, Francesco and Caligiuri, Matteo and Zanuttigh, Pietro},
  booktitle={Proceedings of the IEEE/CVF International Conference on Computer Vision},
  pages={2210--2220},
  year={2023}
}

@inproceedings{zhang2018unreasonable,
  title={The unreasonable effectiveness of deep features as a perceptual metric},
  author={Zhang, Richard and Isola, Phillip and Efros, Alexei A and Shechtman, Eli and Wang, Oliver},
  booktitle={Proceedings of the IEEE conference on computer vision and pattern recognition},
  pages={586--595},
  year={2018}
}

@misc{kingma2017adammethodstochasticoptimization,
      title={Adam: A Method for Stochastic Optimization}, 
      author={Diederik P. Kingma and Jimmy Ba},
      year={2017},
      eprint={1412.6980},
      archivePrefix={arXiv},
      primaryClass={cs.LG},
      url={https://arxiv.org/abs/1412.6980}, 
}

@InProceedings{4dgs,
    author    = {Wu, Guanjun and Yi, Taoran and Fang, Jiemin and Xie, Lingxi and Zhang, Xiaopeng and Wei, Wei and Liu, Wenyu and Tian, Qi and Wang, Xinggang},
    title     = {4D Gaussian Splatting for Real-Time Dynamic Scene Rendering},
    booktitle = {Proceedings of the IEEE/CVF Conference on Computer Vision and Pattern Recognition (CVPR)},
    month     = {June},
    year      = {2024},
    pages     = {20310-20320}
}

@inproceedings{degauss,
      title={Degauss: Dynamic-static decomposition with gaussian splatting for distractor-free 3d reconstruction},
      author={Wang, Rui and Lohmeyer, Quentin and Meboldt, Mirko and Tang, Siyu},
      booktitle={Proceedings of the IEEE/CVF International Conference on Computer Vision},
      pages={6294--6303},
      year={2025}
}

@inproceedings{beziergs,
  title={BézierGS: Dynamic Urban Scene Reconstruction with Bézier Curve Gaussian Splatting},
  author={Ma, Zipei and Jiang, Junzhe and Chen, Yurui and Zhang, Li},
  booktitle={ICCV},
  year={2025},
}

@inproceedings{coda4dgs,
  title={Coda-4dgs: Dynamic gaussian splatting with context and deformation awareness for autonomous driving},
  author={Song, Rui and Liang, Chenwei and Xia, Yan and Zimmer, Walter and Cao, Hu and Caesar, Holger and Festag, Andreas and Knoll, Alois},
  publisher={IEEE/CVF},
  booktitle={IEEE/CVF International Conference on Computer Vision (ICCV)},
  year={2025}
}

@article{shan2023democratizing,
  title={Democratizing photogrammetry: an accuracy perspective},
  author={Shan, Jie and Li, Zhixin and Lercel, Damon and Tissue, Kevan and Hupy, Joseph and Carpenter, Joshua},
  journal={Geo-Spatial Information Science},
  volume={26},
  number={2},
  pages={175--188},
  year={2023},
  publisher={Taylor \& Francis}
}

@inproceedings{zhou2024hugs,
  title={Hugs: Holistic urban 3d scene understanding via gaussian splatting},
  author={Zhou, Hongyu and Shao, Jiahao and Xu, Lu and Bai, Dongfeng and Qiu, Weichao and Liu, Bingbing and Wang, Yue and Geiger, Andreas and Liao, Yiyi},
  booktitle={Proceedings of the IEEE/CVF Conference on Computer Vision and Pattern Recognition},
  pages={21336--21345},
  year={2024}
}

@article{chaturvedi2016integrating,
  title={Integrating dynamic data and sensors with semantic 3D city models in the context of smart cities},
  author={Chaturvedi, Kanishk and Kolbe, Thomas H},
  journal={ISPRS Annals of the Photogrammetry, Remote Sensing and Spatial Information Sciences},
  volume={4},
  pages={31--38},
  year={2016},
  publisher={Copernicus GmbH}
}

@article{dong2018learning,
  title={Learning stratified 3D reconstruction},
  author={Dong, Qiulei and Shu, Mao and Cui, Hainan and Xu, Huarong and Hu, Zhanyi},
  journal={Science China Information Sciences},
  volume={61},
  number={2},
  pages={023101},
  year={2018},
  publisher={Springer}
}

@inproceedings{tancik2022block,
  title={Block-nerf: Scalable large scene neural view synthesis},
  author={Tancik, Matthew and Casser, Vincent and Yan, Xinchen and Pradhan, Sabeek and Mildenhall, Ben and Srinivasan, Pratul P and Barron, Jonathan T and Kretzschmar, Henrik},
  booktitle={Proceedings of the IEEE/CVF conference on computer vision and pattern recognition},
  pages={8248--8258},
  year={2022}
}

@InProceedings{Tang_2025_CVPR,
    author    = {Tang, Jiadong and Gao, Yu and Yang, Dianyi and Yan, Liqi and Yue, Yufeng and Yang, Yi},
    title     = {DroneSplat: 3D Gaussian Splatting for Robust 3D Reconstruction from In-the-Wild Drone Imagery},
    booktitle = {Proceedings of the IEEE/CVF Conference on Computer Vision and Pattern Recognition (CVPR)},
    month     = {June},
    year      = {2025},
    pages     = {833-843}
}

@inproceedings{maxey2024uav,
  title={Uav-sim: Nerf-based synthetic data generation for uav-based perception},
  author={Maxey, Christopher and Choi, Jaehoon and Lee, Hyungtae and Manocha, Dinesh and Kwon, Heesung},
  booktitle={2024 IEEE International Conference on Robotics and Automation (ICRA)},
  pages={5323--5329},
  year={2024},
  organization={IEEE}
}

@inproceedings{yan2024street,
    title={Street Gaussians: Modeling Dynamic Urban Scenes with Gaussian Splatting}, 
    author={Yunzhi Yan and Haotong Lin and Chenxu Zhou and Weijie Wang and Haiyang Sun and Kun Zhan and Xianpeng Lang and Xiaowei Zhou and Sida Peng},
    booktitle={ECCV},
    year={2024}
}

@inproceedings{wang2025vggt,
  title={VGGT: Visual Geometry Grounded Transformer},
  author={Wang, Jianyuan and Chen, Minghao and Karaev, Nikita and Vedaldi, Andrea and Rupprecht, Christian and Novotny, David},
  booktitle={Proceedings of the IEEE/CVF Conference on Computer Vision and Pattern Recognition},
  year={2025}
}

@inproceedings{dust3r_cvpr24,
      title={DUSt3R: Geometric 3D Vision Made Easy}, 
      author={Shuzhe Wang and Vincent Leroy and Yohann Cabon and Boris Chidlovskii and Jerome Revaud},
      booktitle = {CVPR},
      year = {2024}
}

@inproceedings{yao2025uni4d,
  title={Uni4D: Unifying Visual Foundation Models for 4D Modeling from a Single Video},
  author={Yao, David Yifan and Zhai, Albert J and Wang, Shenlong},
  booktitle={Proceedings of the Computer Vision and Pattern Recognition Conference},
  pages={1116--1126},
  year={2025}
}

@article{yang2023deformable3dgs,
    title={Deformable 3D Gaussians for High-Fidelity Monocular Dynamic Scene Reconstruction},
    author={Yang, Ziyi and Gao, Xinyu and Zhou, Wen and Jiao, Shaohui and Zhang, Yuqing and Jin, Xiaogang},
    journal={arXiv preprint arXiv:2309.13101},
    year={2023}
}

@article{zhao2023review,
  title={A review of 3D reconstruction from high-resolution urban satellite images},
  author={Zhao, Li and Wang, Haiyan and Zhu, Yi and Song, Mei},
  journal={International Journal of Remote Sensing},
  volume={44},
  number={2},
  pages={713--748},
  year={2023},
  publisher={Taylor \& Francis}
}

@inproceedings{mildenhall2020nerf,
 title={NeRF: Representing Scenes as Neural Radiance Fields for View Synthesis},
 author={Ben Mildenhall and Pratul P. Srinivasan and Matthew Tancik and Jonathan T. Barron and Ravi Ramamoorthi and Ren Ng},
 year={2020},
 booktitle={ECCV},
}

@inproceedings{martin2021nerf,
  title={Nerf in the wild: Neural radiance fields for unconstrained photo collections},
  author={Martin-Brualla, Ricardo and Radwan, Noha and Sajjadi, Mehdi SM and Barron, Jonathan T and Dosovitskiy, Alexey and Duckworth, Daniel},
  booktitle={Proceedings of the IEEE/CVF conference on computer vision and pattern recognition},
  pages={7210--7219},
  year={2021}
}

@inproceedings{turki2022mega,
  title={Mega-nerf: Scalable construction of large-scale nerfs for virtual fly-throughs},
  author={Turki, Haithem and Ramanan, Deva and Satyanarayanan, Mahadev},
  booktitle={Proceedings of the IEEE/CVF conference on computer vision and pattern recognition},
  pages={12922--12931},
  year={2022}
}

@InProceedings{Lin_2024_CVPR,
    author    = {Lin, Jiaqi and Li, Zhihao and Tang, Xiao and Liu, Jianzhuang and Liu, Shiyong and Liu, Jiayue and Lu, Yangdi and Wu, Xiaofei and Xu, Songcen and Yan, Youliang and Yang, Wenming},
    title     = {VastGaussian: Vast 3D Gaussians for Large Scene Reconstruction},
    booktitle = {Proceedings of the IEEE/CVF Conference on Computer Vision and Pattern Recognition (CVPR)},
    month     = {June},
    year      = {2024},
    pages     = {5166-5175}
}

@inproceedings{
    liu2024citygaussianv2,
    title={CityGaussianV2: Efficient and Geometrically Accurate Reconstruction for Large-Scale Scenes},
    author={Yang Liu and Chuanchen Luo and Zhongkai Mao and Junran Peng and Zhaoxiang Zhang},
    booktitle={The Thirteenth International Conference on Learning Representations},
    year={2025},
    url={https://openreview.net/forum?id=a3ptUbuzbW}
}

@inproceedings{li2021neural,
  title={Neural scene flow fields for space-time view synthesis of dynamic scenes},
  author={Li, Zhengqi and Niklaus, Simon and Snavely, Noah and Wang, Oliver},
  booktitle={Proceedings of the IEEE/CVF conference on computer vision and pattern recognition},
  pages={6498--6508},
  year={2021}
}

@inproceedings{pumarola2021d,
  title={D-nerf: Neural radiance fields for dynamic scenes},
  author={Pumarola, Albert and Corona, Enric and Pons-Moll, Gerard and Moreno-Noguer, Francesc},
  booktitle={Proceedings of the IEEE/CVF conference on computer vision and pattern recognition},
  pages={10318--10327},
  year={2021}
}

@inproceedings{tretschk2021non,
  title={Non-rigid neural radiance fields: Reconstruction and novel view synthesis of a dynamic scene from monocular video},
  author={Tretschk, Edgar and Tewari, Ayush and Golyanik, Vladislav and Zollh{\"o}fer, Michael and Lassner, Christoph and Theobalt, Christian},
  booktitle={Proceedings of the IEEE/CVF international conference on computer vision},
  pages={12959--12970},
  year={2021}
}

@inproceedings{cao2023hexplane,
  title={Hexplane: A fast representation for dynamic scenes},
  author={Cao, Ang and Johnson, Justin},
  booktitle={Proceedings of the IEEE/CVF Conference on Computer Vision and Pattern Recognition},
  pages={130--141},
  year={2023}
}

@inproceedings{fridovich2023k,
  title={K-planes: Explicit radiance fields in space, time, and appearance},
  author={Fridovich-Keil, Sara and Meanti, Giacomo and Warburg, Frederik Rahb{\ae}k and Recht, Benjamin and Kanazawa, Angjoo},
  booktitle={Proceedings of the IEEE/CVF Conference on Computer Vision and Pattern Recognition},
  pages={12479--12488},
  year={2023}
}

@inproceedings{cao2021visdrone,
  title={VisDrone-DET2021: The vision meets drone object detection challenge results},
  author={Cao, Yaru and He, Zhijian and Wang, Lujia and Wang, Wenguan and Yuan, Yixuan and Zhang, Dingwen and Zhang, Jinglin and Zhu, Pengfei and Van Gool, Luc and Han, Junwei and others},
  booktitle={Proceedings of the IEEE/CVF International conference on computer vision},
  pages={2847--2854},
  year={2021}
}

@inproceedings{du2018unmanned,
  title={The unmanned aerial vehicle benchmark: Object detection and tracking},
  author={Du, Dawei and Qi, Yuankai and Yu, Hongyang and Yang, Yifan and Duan, Kaiwen and Li, Guorong and Zhang, Weigang and Huang, Qingming and Tian, Qi},
  booktitle={Proceedings of the European conference on computer vision (ECCV)},
  pages={370--386},
  year={2018}
}

@inproceedings{barekatain2017okutama,
  title={Okutama-action: An aerial view video dataset for concurrent human action detection},
  author={Barekatain, Mohammadamin and Mart{\'\i}, Miquel and Shih, Hsueh-Fu and Murray, Samuel and Nakayama, Kotaro and Matsuo, Yutaka and Prendinger, Helmut},
  booktitle={Proceedings of the IEEE conference on computer vision and pattern recognition workshops},
  pages={28--35},
  year={2017}
}

@article{chen2022stpls3d,
  title={Stpls3d: A large-scale synthetic and real aerial photogrammetry 3d point cloud dataset},
  author={Chen, Meida and Hu, Qingyong and Yu, Zifan and Thomas, Hugues and Feng, Andrew and Hou, Yu and McCullough, Kyle and Ren, Fengbo and Soibelman, Lucio},
  journal={arXiv preprint arXiv:2203.09065},
  year={2022}
}

@article{wang2025uavscenes,
  title={UAVScenes: A Multi-Modal Dataset for UAVs},
  author={Wang, Sijie and Li, Siqi and Zhang, Yawei and Yu, Shangshu and Yuan, Shenghai and She, Rui and Guo, Quanjiang and Zheng, JinXuan and Howe, Ong Kang and Chandra, Leonrich and others},
  journal={arXiv preprint arXiv:2507.22412},
  year={2025}
}

@article{choi2025uav4d,
  title={Uav4d: Dynamic neural rendering of human-centric uav imagery using gaussian splatting},
  author={Choi, Jaehoon and Jung, Dongki and Maxey, Christopher and Lee, Yonghan and Eum, Sungmin and Manocha, Dinesh and Kwon, Heesung},
  journal={arXiv preprint arXiv:2506.05011},
  year={2025}
}

@InProceedings{karaev2024cotracker3,
    author    = {Nikita Karaev and Iurii Makarov and Jianyuan Wang and Natalia Neverova and Andrea Vedaldi and Christian Rupprecht},
    title     = {{CoTracker3}: Simpler and Better Point Tracking by Pseudo-Labelling Real Videos},
    journal   = {arxiv},
    year      = {2024}
  }
}


\end{document}